\documentclass[lettersize,journal]{IEEEtran}
\usepackage{amsmath,amsfonts}
\usepackage{algorithmic}
\usepackage{algorithm}
\usepackage{array}
\usepackage{multirow}
\usepackage{footmisc}
\usepackage[caption=false,font=normalsize,labelfont=sf,textfont=sf]{subfig}
\usepackage{textcomp}
\usepackage{stfloats}
\usepackage{url}
\usepackage{verbatim}
\usepackage{graphicx}
\usepackage{cite}
\usepackage{float}
\usepackage{tikz,xcolor,hyperref}
\hypersetup{hidelinks,
	colorlinks=true,
	allcolors=black,
	pdfstartview=Fit,
	breaklinks=true}

\definecolor{lime}{HTML}{A6CE39}
\DeclareRobustCommand{\orcidicon}{
\begin{tikzpicture}
\draw[lime, fill=lime] (0,0)
circle[radius=0.16]
node[white]{{\fontfamily{qag}\selectfont \tiny \.{I}D}};
\end{tikzpicture}
\hspace{-2mm}
}
\foreach \x in {A, ..., Z}{%
\expandafter\xdef\csname orcid\x\endcsname{\noexpand\href{https://orcid.org/\csname orcidauthor\x\endcsname}{\noexpand\orcidicon}}
}

\begin{document}

\title{WHU-Stereo: A Challenging Benchmark for Stereo Matching of High-Resolution Satellite Images}

\newcommand{\orcidauthorA}{0000-0002-0436-2768}
\newcommand{\orcidauthorB}{0000-0001-6754-2865}
\newcommand{\orcidauthorC}{0000-0002-7799-650X}
\newcommand{\orcidauthorD}{0000-0002-3162-0566}
\newcommand{\orcidauthorE}{0000-0002-5565-1033}

\author{Shenhong~Li\hspace{-1.5mm}\orcidA{},~Sheng~He\hspace{-1.5mm}\orcidB{},~San~Jiang\hspace{-1.5mm}\orcidC{},~Wanshou~Jiang\hspace{-1.5mm}\orcidD{},~and~Lin~Zhang\hspace{-1.5mm}\orcidE{}
\thanks{S. Li, S. He, and W. J are with the State Key Laboratory of Information Engineering in Surveying, Mapping and Remote Sensing, Wuhan University, Wuhan, 430072, China (e-mail: shenhonglee@whu.edu.cn; 2014301610342@whu.edu.cn; jws@whu.edu.cn).  S. Li and S. He made equal contributions to this paper.}
\thanks{S. Jiang is with the School of Computer Science, China University of Geosciences, Wuhan, 430074, China (e-mail: jiangsan@cug.edu.cn).}
\thanks{L. Zhang is with the School of Resource and Enviornmental Sciences, Wuhan University, Wuhan, 430072, China (e-mail: lynnzhang@whu.edu.cn).}
\thanks{Corresponding author: Wanshou Jiang.}}



\maketitle

\begin{abstract}
Stereo matching of high-resolution satellite images (HRSI) is still a fundamental but challenging task in the field of photogrammetry and remote sensing. Recently, deep learning (DL) methods, especially convolutional neural networks (CNNs), have demonstrated tremendous potential for stereo matching on public benchmark datasets. Similar to other DL applications, mainstream stereo matching models depend on huge quantities of training data with ground-truth labels for parameter learning. However, datasets for stereo matching of satellite images are scarce, which profoundly blocks the research and usage of DL techniques in this field. To facilitate further research, this paper creates and publishes a challenging dataset, termed WHU-Stereo, for stereo matching DL network training and testing. This dataset is created by using airborne LiDAR point clouds and high-resolution stereo imageries taken from the Chinese GaoFen-7 satellite (GF-7). For generating ground-truth disparities, occlusions should be seriously considered since they usually cause difficulty and degenerate precision of labels. To address this issue, an occlusion removal technique is proposed in this study, which can adapt to point clouds with different densities and shows high potential in training data preparation. The WHU-Stereo dataset contains more than 1700 epipolar rectified image pairs, which cover six areas in China and includes various kinds of landscapes. We have assessed the accuracy of ground-truth disparity maps, and it is proved that our dataset achieves comparable precision compared with existing state-of-the-art stereo matching datasets. To verify its feasibility, in experiments, the hand-crafted SGM stereo matching algorithm and recent deep learning networks have been tested on the WHU-Stereo dataset. Experimental results show that deep learning networks can be well trained and achieves higher performance than hand-crafted SGM algorithm, and the dataset has great potential in remote sensing application. The WHU-Stereo dataset can serve as a challenging benchmark for stereo matching of high-resolution satellite images, and performance evaluation of deep learning models. Our dataset is available at \url{https://github.com/Sheng029/WHU-Stereo}.
\end{abstract}


\section{Introduction}
\IEEEPARstart{A}{s} one of the most active research areas in photogrammetry and remote sensing, stereo matching of high-resolution satellite images plays an important role in the large-scale 3D reconstruction of the earth's surface. The purpose of stereo matching is to obtain accurate correspondence points for each pixel and generate a disparity map for epipolar rectified stereo images\cite{ref1,ref2}, which is further used to recover the depth and 3D information. Stereo matching has a wide range of applications, including the reconstruction of city-scale 3D building models or even country-scale scenes with an impressive level of detail\cite{ref3}.

Stereo matching has been extensively studied for decades. Generally, the stereo matching pipeline consists of four steps: matching cost computation, cost aggregation, disparity computation and optimization, and disparities refinement\cite{ref4}. Existing methods can be classified into two categories, including hand-crafted methods and learning-based methods. Hand-crafted methods have made significant progress and been widely used, e.g., GraphCut\cite{ref5}, semi-global matching (SGM)\cite{ref6}, PatchMatch\cite{ref7}, and so on. However, detrimental flaws still exist. For example, due to the algorithm complexity being high and all the feature engineering being empirically designed, they usually lack efficiency as well as robustness and perform less than satisfactory in ill-posed regions like texture-less areas, repetitive patterns, and occlusions. These shortcomings make it difficult to meet the requirements of processing large-volume complex remote sensing images.

In recent years, deep learning technologies, especially convolutional neural networks (CNNs), have been extensively used for various vision tasks\cite{ref8,ref9,ref10,ref11} and have achieved great success. The rapid development of deep learning has also benefited the research of stereo matching. Early studies tend to apply CNNs to learn parts of the standard four-step pipeline to improve the individual steps. For example, \cite{ref12} proposes MC-CNN that automatically learns matching costs via a Siamese CNN structure. \cite{ref13} proposes SGM-Net that introduces a CNN learning penalty term in the standard process of SGM. \cite{ref14} proposes LRCR that embeds the left-right consistency check into a unified pipeline to improve the final disparity estimate. Although these methods have substantially improved some steps of the traditional pipeline, the introduction of end-to-end models drives the community towards a new paradigm\cite{ref15} by integrating the established pipeline\cite{ref4} into a single network and directly predicting disparity maps from input stereo pairs. Since the release of DispNet\cite{ref16} and GC-Net\cite{ref17}, state-of-the-art end-to-end models have been rapidly developed. Their matching performance obtained on benchmarks\cite{ref16,ref18,ref19,ref20,ref21} has exceeded traditional methods in both speed and accuracy. For example, PSM-Net\cite{ref22} is a pyramid stereo matching network that consists of spatial pyramid pools (SPP) and 3D stacked hourglass modules. It combines global context into stereo matching to achieve a reliable estimation even with occlusions, texture-less areas, or repeated patterns. It ranked first in the KITTI 2012\cite{ref18} and 2015\cite{ref19} leaderboards at that time. To improve efficiency, Stereo-Net\cite{ref23} is designed as a lightweight network, which generates a coarse disparity map at a low-resolution scale and gradually restores it to the original resolution with the guidance of input images. Stereo-Net is the first end-to-end model that achieves real-time stereo matching.

Since deep learning-based methods have dominated stereo matching of close-range images\cite{ref16,ref18,ref19,ref20,ref21}, it is natural to consider applying them to remote sensing images. However, fewer network models have been wholly designed for stereo matching of remote sensing images. As far as we know, no published papers have systematically studied this topic before 2019. \cite{ref1} first investigated the application of deep learning-based stereo methods on aerial remote sensing images and compared them with conventional methods. The authors created an aerial dataset and used it to train and test MC-CNN\cite{ref12}, DispNet\cite{ref16}, and GC-Net\cite{ref17}. However, the dataset has not been published, and too limited impact has been brought because such a private dataset cannot ensure comparing studies and may hinder improving algorithms\cite{ref24}. Until the Data Fusion Contest 2019, \cite{ref25} introduced the Urban Semantic 3D (US3D) dataset\cite{ref26} that promoted the design of deep learning models for stereo matching of remote sensing images. For example, \cite{ref27} proposed a multi-task architecture named bidirectional guided attention network (BGA-Net) that can jointly conduct semantic segmentation and disparity estimation, it set a state of the art on the pairwise semantic stereo challenge in the contest. Based on PSM-Net\cite{ref22}, \cite{ref28} developed an edge-sense bidirectional pyramid stereo matching network (Bidir-EPNet) that embedded a forward-backward consistency assumption, it improved disparity estimation of occluded and texture-less areas. For high efficiency, \cite{ref29} adopted factorized convolutions to design a lightweight dual-scale matching network (DSM-Net) and achieved an excellent tradeoff between computation and accuracy. To handle satellite image pairs with complex scenes and large disparity range, \cite{ref30} introduced a hierarchical multi-scale matching network (HMSM-Net) which hierarchically learns stereo correspondence at multiple scales and achieved an impressive performance.

Although the US3D dataset inspired a few researchers, the application of deep learning for stereo matching of remote sensing images has not attracted much attention. As the current deep learning is data-driven, the accuracy of deep learning models depends heavily on the training data set\cite{ref24}. Open source datasets have dramatically promoted the development of deep learning methods for other tasks, such as semantic segmentation of remote sensing images\cite{ref24,ref31,ref32}. However, large-scale, high-quality datasets generated from aerial or satellite stereo imagery are scarce. As a result, researchers have to spend a lot of time on finding and constructing datasets, which heavily blocks the study and usage of deep learning on stereo matching in the field of photogrammetry and remote sensing. To promote further the research of stereo matching for remote sensing images, especially satellite imagery, this study has created and published a challenging training and testing dataset for stereo matching of high-resolution satellite images.

In the literature, apart from the US3D dataset, there only exist two open-source remote sensing datasets for stereo matching. For aerial photogrammetry, \cite{ref33} proposed a dataset based on the ISPRS 3D reconstruction benchmark\cite{ref34}. Instead of using digital surface models (DSM) to generate ground-truth disparities, \cite{ref33} used LiDAR point clouds to generate more accurate labels. However, the disparity density is too sparse. \cite{ref35} proposed the SatStereo dataset that was prepared by using images from WorldView-3 (a minority from WorldView-2). Except for dense disparity labels, it also provides a building mask and metadata for each image. US3D\cite{ref26} is the largest dataset with high-quality dense labels. However, it only covers two cities, and many image pairs are captured from the same districts. Besides, images of the SateStereo and US3D are collected at different times. Thus, the land cover of an image pair is likely to show seasonal appearance differences. To our knowledge, SatStereo and US3D datasets are the only two datasets that include stereo-rectified image pairs and ground-truth disparities generated from satellite imagery and LiDAR.

To affiliate the lack of training datasets, we manually edit a satellite image dataset of stereo samples and name it WHU-Stereo. Table \ref{tab1} lists the comparisons between these datasets and ours. The images are collected from the Chinese GF-7 satellite that carries a stereo camera, which ensures that all image pairs are captured at the same time. To generate ground-truth disparities of high accuracy, we use airborne LiDAR data. Accordingly, we propose a pipeline for disparity generation using LiDAR point clouds. This pipeline incorporates an occlusion removal technique that can handle difficulties caused by occlusions and guarantee high-quality ground truths. Besides, we have assessed the accuracy of the ground-truth disparity maps, and it is proved that our dataset is comparable to the same in the existing state-of-the-art stereo datasets. Our dataset covers six areas in China and includes various districts and landscapes. We have used it to evaluate some typical stereo matching methods, including a traditional algorithm and three deep learning-based models. Experiments show that this dataset can serve as a challenging benchmark and is suitable for testing the generalization and extrapolation ability of deep learning methods.

\begin{table}[!h]
\caption{Comparison among different datasets for stereo matching of remote sensing images (only stereo pairs that have publicly available ground-truth disparity labels are counted here)}
\centering
\begin{tabular}{cccccc}
\hline
\textbf{Dataset} & \textbf{Year} & \textbf{Scene} & \textbf{Mode} & \textbf{Number} & \textbf{Density} \\ [2pt]
\hline
\cite{ref33} & 2021 & Aerial & IRRG & 1092 & Sparse \\ [2pt]
SatStereo\cite{ref35} & 2019 & Satellite & Panchromatic & 72 & Dense \\ [2pt]
US3D\cite{ref26} & 2019 & Satellite & RGB & 4292 & Dense \\ [2pt]
\textbf{WHU-Stereo} & 2022 & Satellite & Panchromatic & 1757 & Dense \\ [2pt]
\hline
\end{tabular}
\label{tab1}
\end{table}

The main contributions of this paper are:

\begin{itemize}
\item We provide WHU-Stereo, a large, accurate, and open-source dataset for stereo matching of high-resolution satellite images. As far as we know, it is the first stereo benchmark that uses images from the Chinese GF-7 satellite.
\item We propose a pipeline incorporating an occlusion removal technique for generating ground-truth disparities from LiDAR point clouds. This feasible and universal pipeline can adapt to point clouds with different densities and reduce the impact of occlusions.
\item We thoroughly evaluate a conventional method and several recent deep learning methods on the same benchmark and the results can sever as the baseline of this benchmark, which may attract more attention to this field.
\end{itemize}

The following sections are arranged as follows. Section II gives a brief description of the data source. Section III describes how the dataset is generated in detail. In Section IV, the quality of this dataset is qualitatively and quantitatively evaluated. In Section V, experiments are conducted to evaluate stereo matching methods using this dataset, meanwhile, a baseline is set for this dataset. A conclusion and further prospects are made in Section VI.

\section{Satellite Images and LiDAR Point Clouds}
\subsection{Test Sites}
\begin{figure*}[!t]
\centering
\includegraphics[width=6.5in]{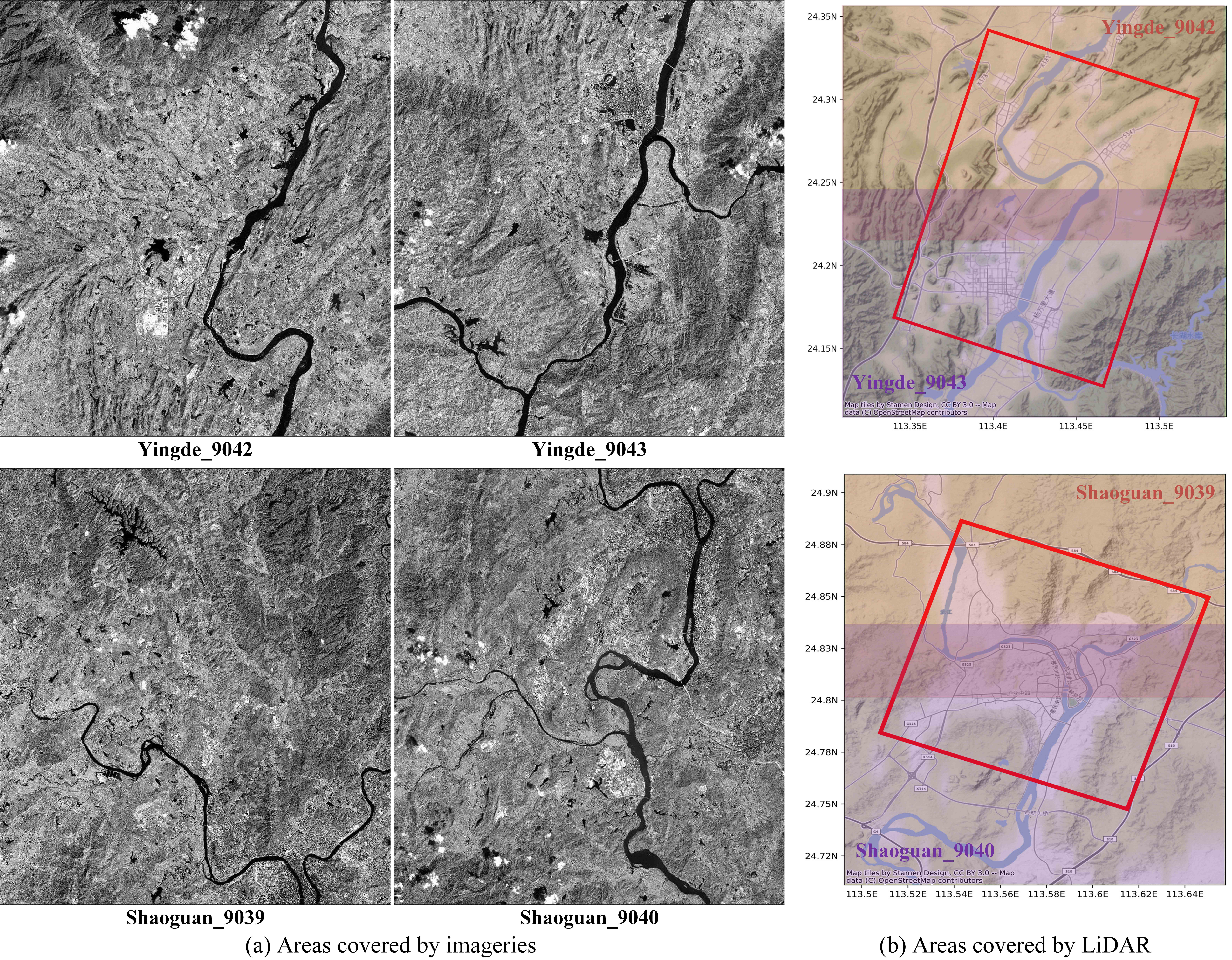}
\caption{GF-7 satellite imageries and corresponding LiDAR data coverage. (a) Imageries are captured over the cities of Yingde and Shaoguan (only the imageries of backward view are displayed here). (b) The approximate coverage of LiDAR data is shown in OpenStreetMap.}
\label{fig1}
\end{figure*}

Current datasets\cite{ref26,ref33,ref35} are collected from cities in Europe and America, our dataset will be a beneficial supplement as it is the first open-source stereo dataset containing images covering Chinese cities that have dramatically different geographical features compared to the formers. WHU-Stereo adopts satellite imageries captured from six cities, namely Wuhan, Hengyang, Shaoguan, Kunming, Yingde, and Qichun. Due to these cities being located in areas of different terrain (e.g., plain, plateau, mountain, and river) and at different stages of development, they have various land covers and can roughly represent cities of China. The satellite imageries cover approximately 900 square kilometers, which is a big range. Fig. \ref{fig1} lists the imagery coverage of two cities (as well as the corresponding airborne LiDAR coverage). It can be seen that the terrain is undulating and covered with a variety of landscapes such as vegetation, water, urban regions, and so on. The complex environment makes the dataset an ideal benchmark to evaluate the performance and potential of stereo matching algorithms.

\subsection{Satellite Imagery}
The China Centre for Resources Satellite Data and Application provided us with GF-7 imageries for our stereo image pairs. GF-7 is the first satellite designed for 3D mapping in China. Table \ref{tab2} summarizes the main technical parameters of the optical sensor carried by GF-7. It is equipped with a dual-line array stereoscopic camera, which can effectively acquire panchromatic stereo imageries with a width larger than 20 kilometers and a spatial resolution better than 80 centimeters, as well as multispectral imageries with a spatial resolution of 2.6 meters. The backward-view camera has an inclination of -5 degrees while that of the forward-view camera is 26 degrees, enabling it to capture sufficient information on the earth's surface and meet the requirement of large-scale stereo mapping. The radiometric resolution of the camera is 11 bits and imageries are stored in a format of 16-bit. These characteristics make GF-7 satellite imageries suitable for producing stereo matching datasets.

\begin{table}[!h]
\caption{Main technical parameters of the optical sensor carried by the GF-7 satellite}
\centering
\begin{tabular}{c|c|c}
\hline
\textbf{Item} & \multicolumn{2}{c}{\textbf{Parameter}} \\ [2pt]
\hline
\multirow{5}{*}{Spectrum} & Panchromatic & 0.45$\sim$0.9 $\mu m$ \\ [2pt]
\cline{2-3}
	& \multirow{4}{*}{Multispectral}  & 0.45$\sim$0.52 $\mu m$ \\ [2pt]
	&                                                  & 0.52$\sim$0.59 $\mu m$ \\ [2pt]
	&                                                  & 0.63$\sim$0.69 $\mu m$ \\ [2pt]
	&                                                  & 0.77$\sim$0.89 $\mu m$ \\ [2pt]
\hline
Forward view & \multicolumn{2}{c}{$+26^{\circ}$}     \\ [2pt]
Backward view & \multicolumn{2}{c}{$-5^{\circ}$}   \\ [2pt]
\hline
\multirow{2}{*}{Spatial resolution} & Panchromatic & Forward: 0.8m, backward: 0.65m  \\ [2pt]
	& Multispectral & Backward: 2.6m \\ [2pt]
\hline
Width & \multicolumn{2}{c}{$\geq$20km} \\ [2pt]
\hline
\end{tabular}
\label{tab2}
\end{table}

Our dataset collected nine GaoFen-7 panchromatic imageries over Kunming, Qichun, Wuhan, Yingde, and Shaoguan in 2020 and two GaoFen-7 panchromatic imageries between 2020 and 2021 over Hengyang. All images are in Tag Image File Format (TIFF) with Rational Polynomial Coefficient (RPC) sensor model metadata.

\subsection{Airborne LiDAR}
\begin{figure}[!t]
\centering
\includegraphics[width=3.45in]{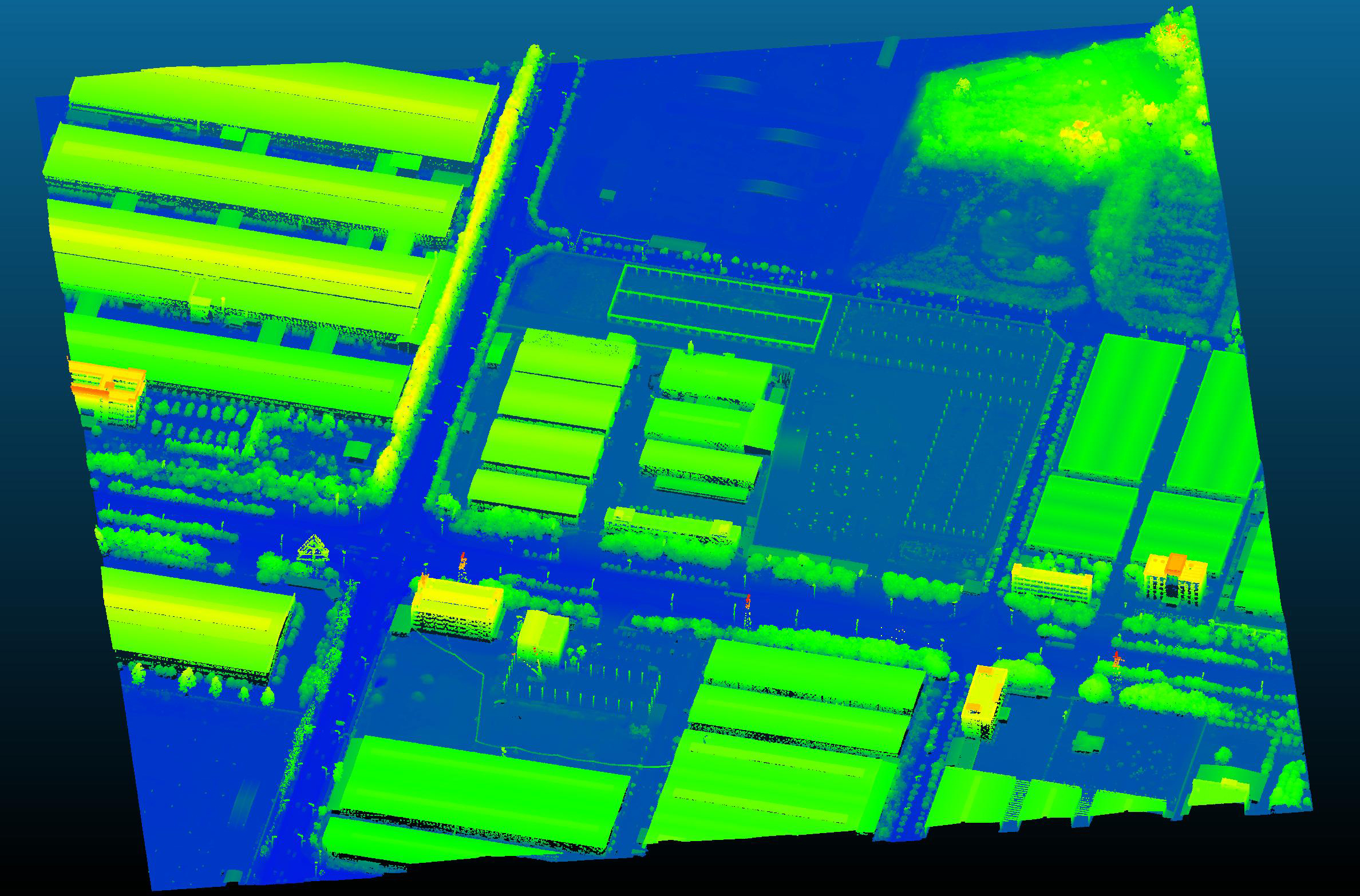}
\caption{An example of LiDAR point cloud (visualized by the CloudCompare software).}
\label{fig2}
\end{figure}

Ground-truth geometry for stereo evaluation is provided by airborne LiDAR collected over the six areas by a third-party company. The LiDAR aggregate nominal pulse spacing (ANPS) is approximately 25cm, roughly equivalent to a third or quarter of a pixel in the satellite imagery. An example of the LiDAR point clouds is shown in Fig. \ref{fig2}.

\section{Dataset Generation}
\begin{figure*}[!t]
\centering
\includegraphics[width=6.5in]{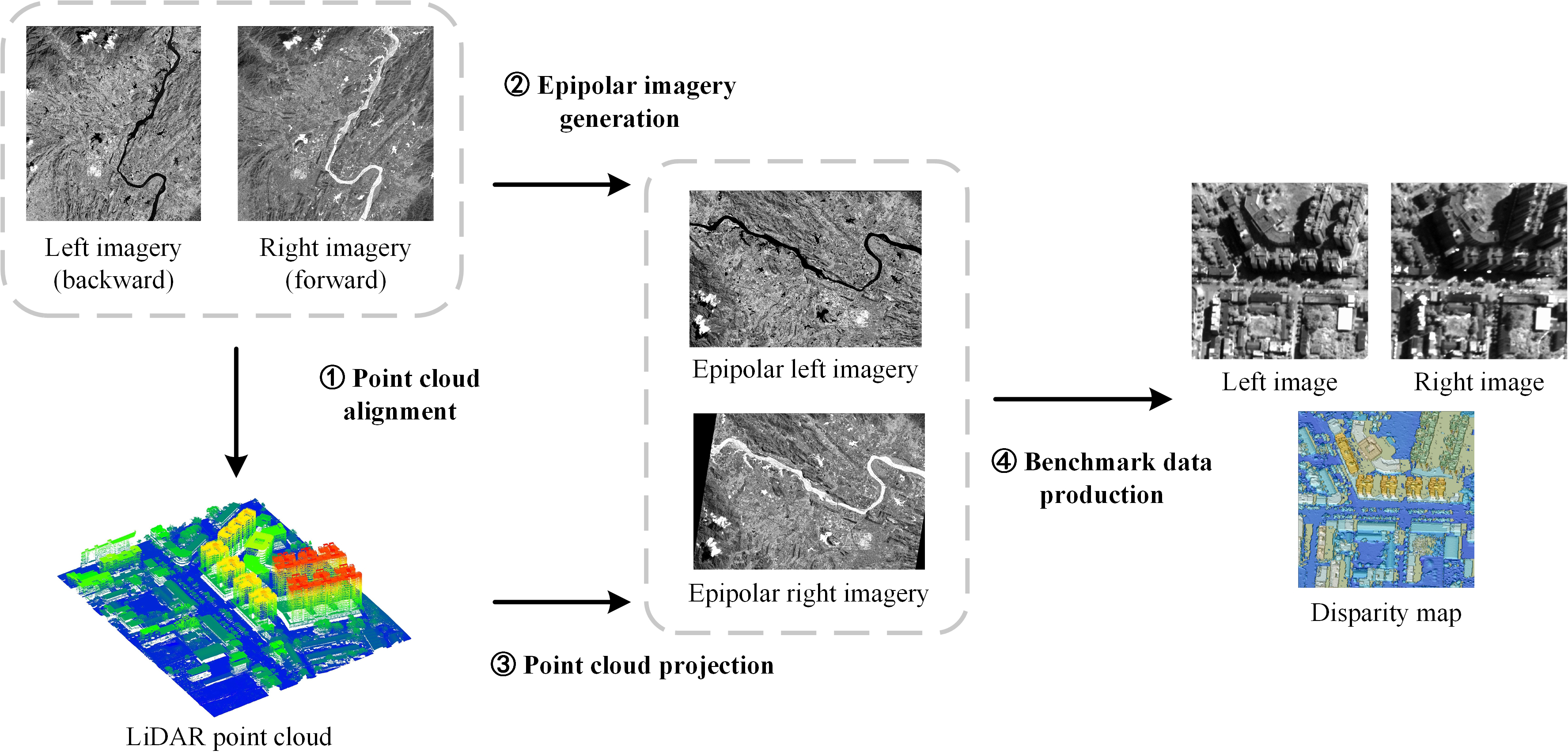}
\caption{The overall pipeline of data generation. Giving a pair of original satellite imageries and corresponding cloud points, a set of epipolar rectified image pairs and disparity maps are generated after the four steps.}
\label{fig3}
\end{figure*}

The dataset is composed of epipolar rectified image pairs and ground-truth disparity maps, which are generated by four steps: point cloud alignment, epipolar imagery generation, point cloud projection, and benchmark data production. The overall pipeline is shown in Fig. \ref{fig3}.

\subsection{Point Cloud Alignment}
\begin{figure}[!t]
\centering
\includegraphics[width=3.45in]{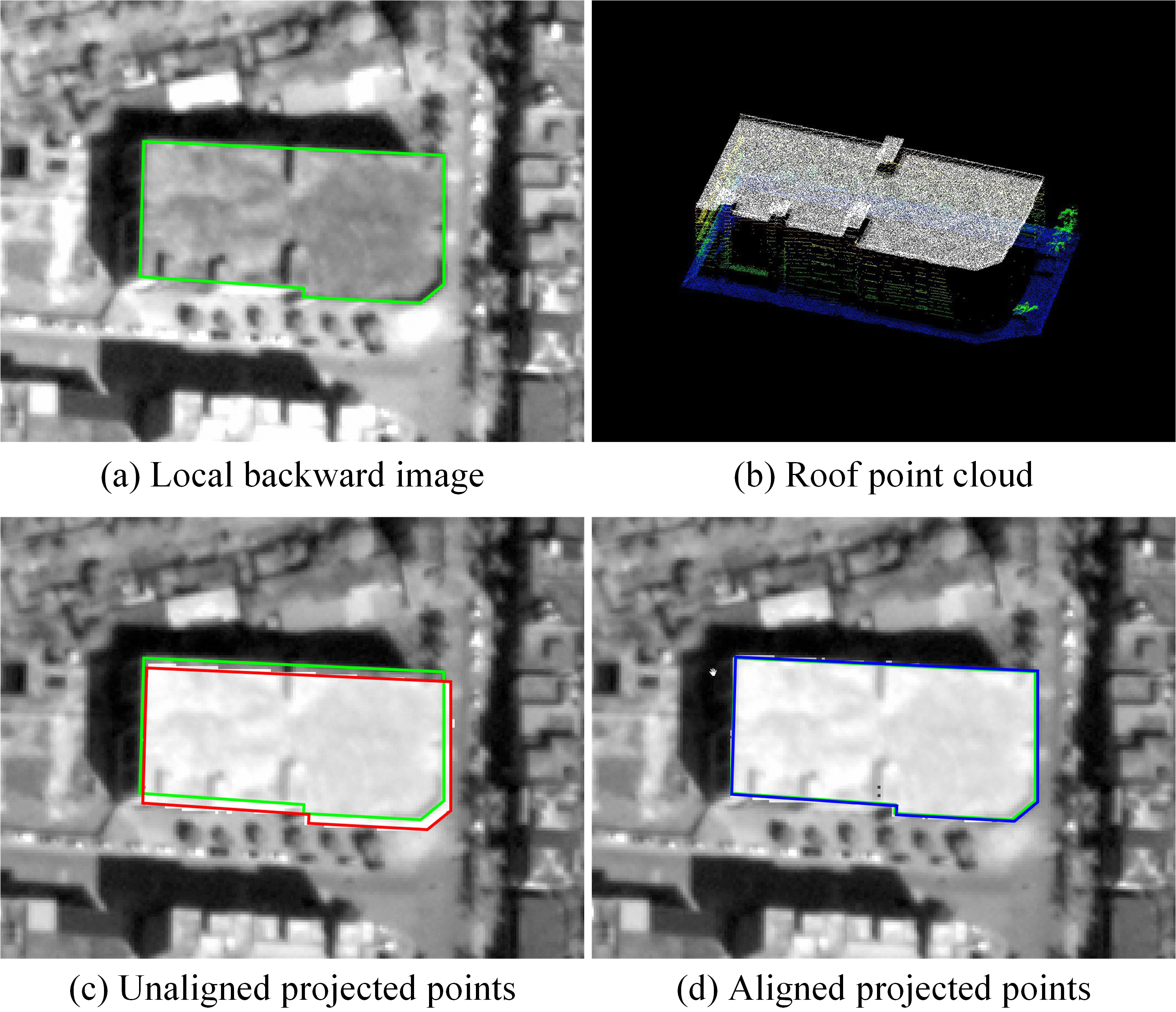}
\caption{Comparison of projected points before and after alignment. The green outline is the ground truth of the roof contour. The red and blue outlines are the contours of unaligned and aligned projected points, respectively.}
\label{fig4}
\end{figure}

The dataset generation pipeline begins with aligning LiDAR point clouds on satellite imageries to compensate for errors of RPC models. If an RPC model was directly used for a point cloud, the projected coordinates would deviate from the imagery. We take an affine transform to project points on the imagery to reduce the offset. The affine transformation parameters are calculated by manually selecting multiple ground control points on the imagery. We choose the sharp corners of buildings as ground control points. These ground control points are distributed evenly around each area. Affine transformation can be achieved by many third libraries, e.g., Eigen, OpenCV, and boost. Since ground control points are two-dimensional, we prefer to choose OpenCV to calculate the affine transformation parameters. Fig. \ref{fig4} shows a comparison of the roof projected points before and after alignment. An offset exists between the roof contour of the unaligned projected points and the ground truth. After applying the affine transformation, the aligned projected points reduce the offset.

\subsection{Epipolar Imagery Generation}
This procedure aims to generate epipolar rectified stereo imageries, in which vertical disparities are eliminated and disparities only exist in the horizontal direction. This can be conveniently achieved with our open-source software, OpenRS\footnote{Details and download link of the software can be found at \url{http://openrs.whu.edu.cn/download.html}.}. The backward imagery is set as the left imagery (also called the reference imagery) and the forward imagery is set as the right imagery (also called the target imagery). We first compute the direction of the epipolar line, then rectify the imagery pair. Finally, the corresponding orientation parameter files are generated to transform aligned projected point clouds into rectified imageries.

\subsection{Point Cloud Projection}
At this point, we compute the ground-truth disparity maps from the LiDAR point cloud. Our pipeline uses RPC models and affine transform to project and align the LiDAR point cloud into the imagery plane. It then transforms the aligned projected point cloud into rectified left and right imageries using the orientation parameters. The disparity $d$ is defined as:
\begin{equation}
d=x_{l}-x_{r}
\end{equation}
where $x_{l}$ and $x_{r}$ are the projection coordinates of a 3D point on the x-axis of the left and right epipolar imageries, respectively.

\begin{figure}[!t]
\centering
\includegraphics[width=3.45in]{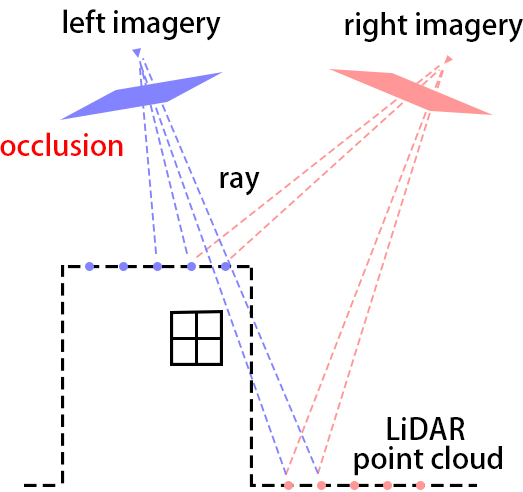}
\caption{LiDAR point cloud occluded in an image.}
\label{fig5}
\end{figure}

\begin{figure}[!h]
\centering
\includegraphics[width=3.45in]{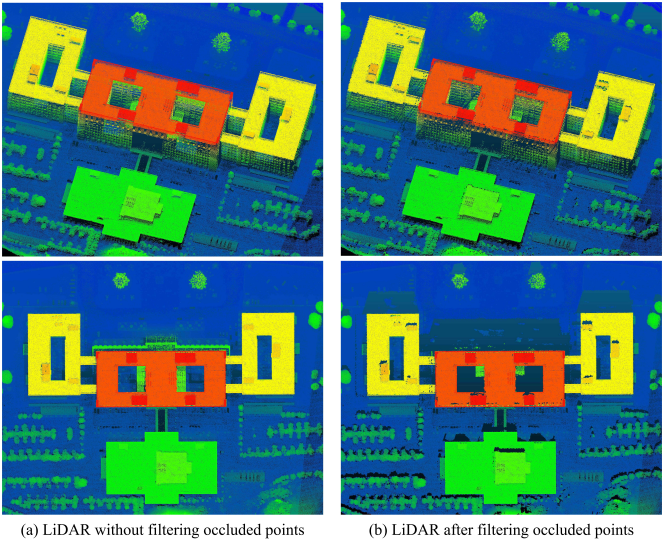}
\caption{Comparison of point cloud before and after removing occluded points.}
\label{fig6}
\end{figure}

Because the point cloud is sparse, it is difficult to predict the occlusions which can be important\cite{ref36}. Fig. \ref{fig5} shows an example of a point cloud with occlusions in the left imagery, some points from the ground and building façade should be removed as the roof occludes them. Though researchers have proposed density-based\cite{ref33} and rasterization-based\cite{ref37} filtering methods, the variation in density makes it difficult to determine the filtering threshold, and rasterization may result in data loss and accuracy falling. We propose a filter based on the correlation coefficient to remove occluded points. The filter firstly projects a point cloud on the left and right epipolar imageries to generate homologous points. Then, it creates a $9\times9$ window centered on each projected point. Next, the correlation coefficient between the windows is calculated to determine whether the homologous point is in the same region. Finally, the point will be removed if the correlation coefficient is less than a threshold. This filter removes occluded points effectively and can adapt to point clouds with different densities, which is feasible and universal. An example of a point cloud in which occluded points have been removed is shown in Fig. \ref{fig6}(b).

The density of the point cloud is larger than the image resolution. Each pixel has multiple non-occluded projected points $\left\{p_{i_{1}}, p_{i_{2}}, \ldots, p_{i_{k}}\right\}$. A pixel corresponding to the 3D space coordinate is expressed as $p_{i}\left(\overline{\sum X_{l}}, \overline{\sum Y_{l}}, \max \left(Z_{i}\right)\right)$, where $\left(\overline{\sum X_{l}}, \overline{\sum Y_{l}}\right)$ and $\max \left(Z_{i}\right)$ are the average $(X, Y)$ coordinate and the maximum $Z$ coordinate of all points that the pixel has, respectively. A 3D point $p_{i}$ that is related to the pixel $\left(x_{l}, y\right)$ of left imagery is projected onto the right imagery to locate the homologous pixel $\left(x_{r}, y\right)$ We can apply the definition of disparity to get a disparity value at each point.

\subsection{Benchmark Data Generation}
\begin{figure}[!t]
\centering
\includegraphics[width=3.45in]{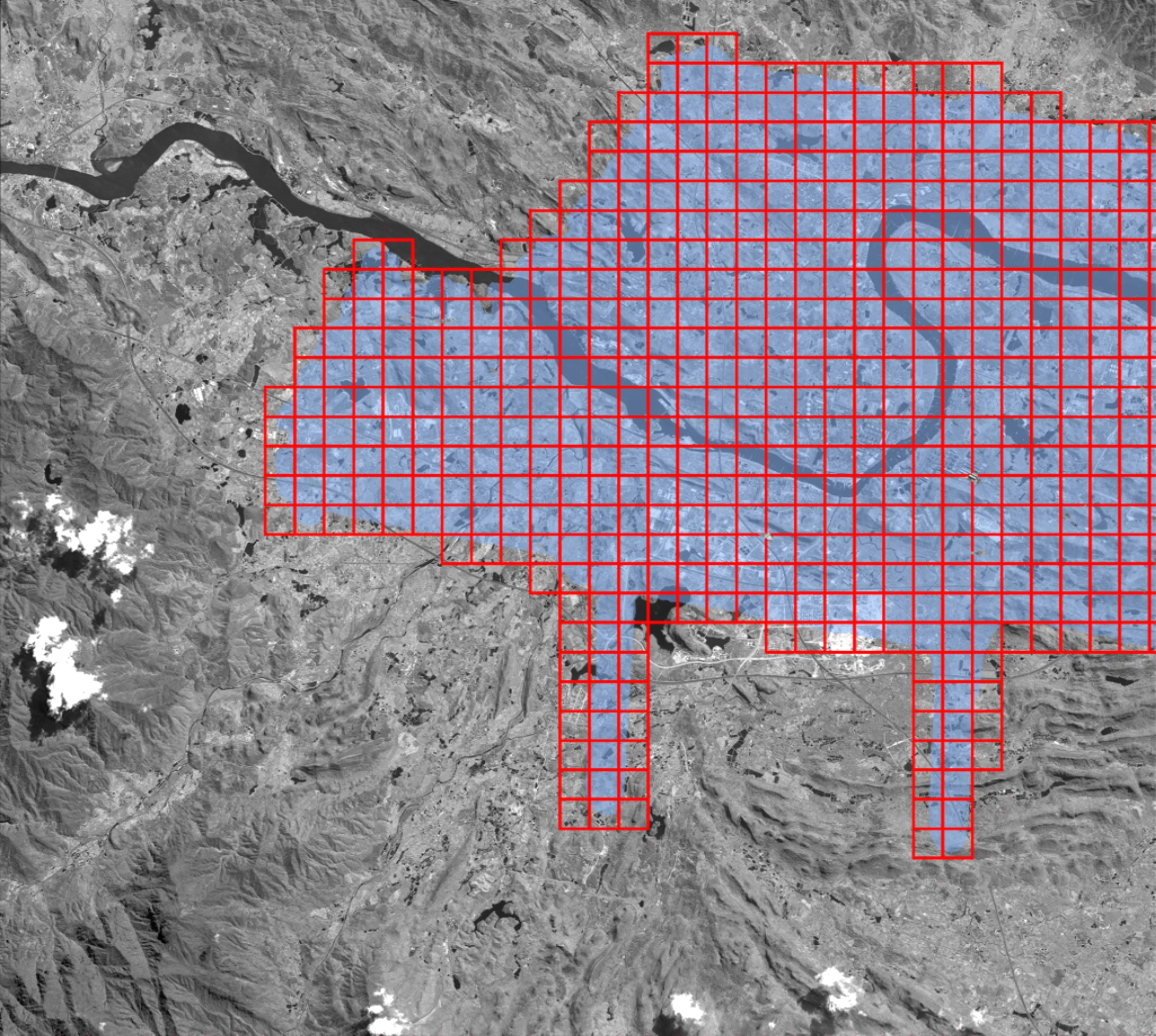}
\caption{Cropped images in the Yingde area. The red grids are the cropped images. The blue area is the coverage region of the point cloud.}
\label{fig7}
\end{figure}

After obtaining epipolar rectified imageries and disparity maps in the previous steps, we produce the benchmark data. To avoid extremely large disparity values, we set the offset of the right imagery to the average disparity in this area. Due to the size of original imageries and disparity maps being large, they have to be cropped to fit in with the computer memory. Similar to the US3D dataset, the final data are $1024\times1024$ cropped images with a 16-bit depth panchromatic band, and the disparity maps are stored on 16-bit float values. Because point clouds distribute irregularly on the images, we need to remove the cropped images with incomplete disparity maps. Fig. \ref{fig7} shows the cropped images in Yingde City. Since the incomplete coverage of the point cloud, part of the cropped images located at the boundary needs to be removed.

\begin{figure*}[!t]
\centering
\includegraphics[width=7in]{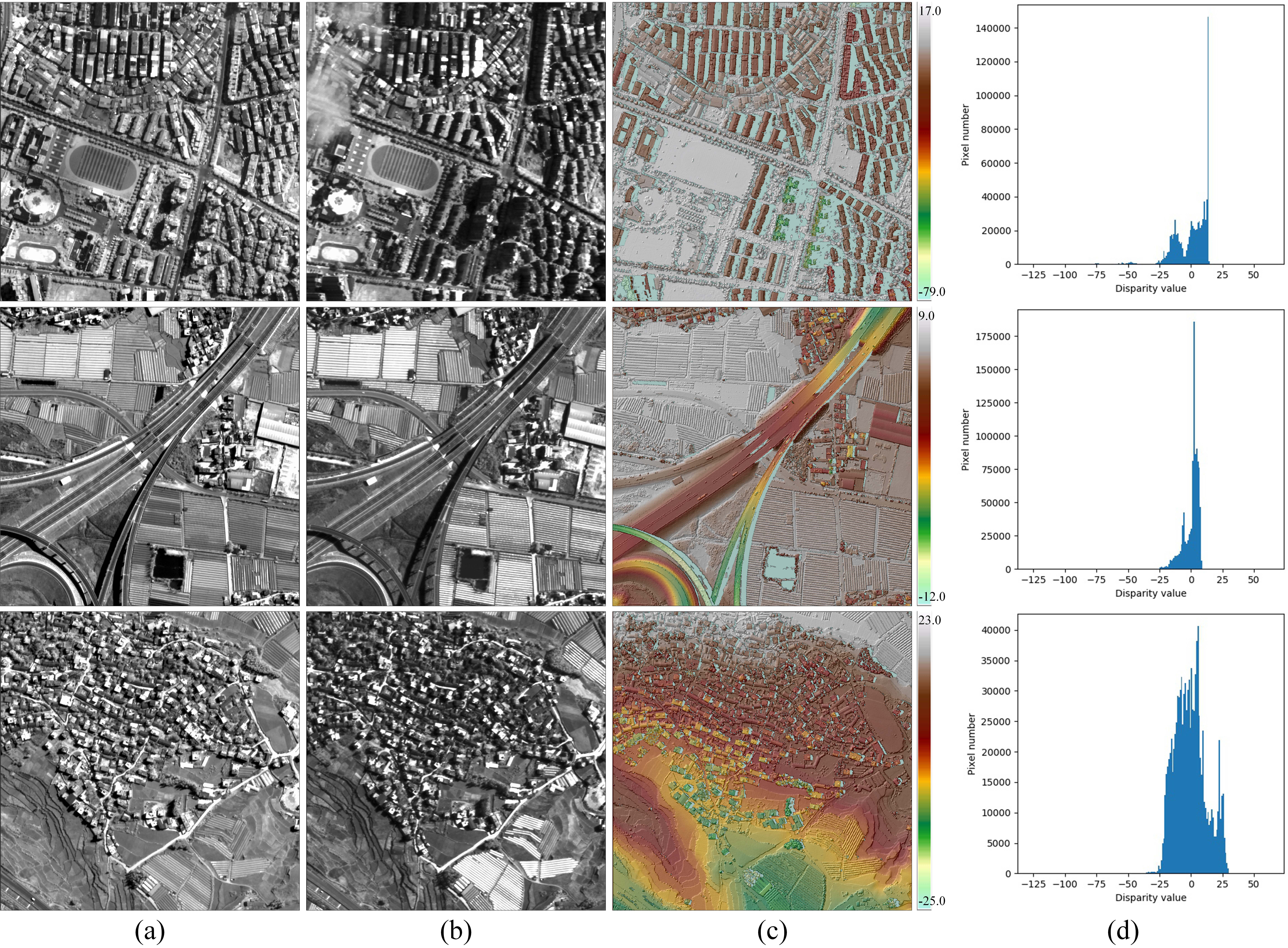}
\caption{Examples of epipolar rectified image pairs and corresponding disparity maps. (a) Left images. (b) Right images. (c) Disparity maps (best viewed in color). (d) Distributions of disparity values. The visualization is performed using OpenRS.}
\label{fig8}
\end{figure*}

After the benchmark data generation, we obtain 1981 pairs of epipolar rectified stereo images, among which 1757 pairs have ground-truth labels and the remainder doesn’t. Several samples are shown in Fig. \ref{fig8}. We split the dataset into three subsets, namely training, validation, and testing sets, as depicted in Table \ref{tab3}. Among the six cities, Shaoguan, Kunming, Yingde, and Qichun are used for evaluating deep learning models’ ability of geographical generalization within a city. Wuhan and Hengyang are used for evaluating the ability of geographical extrapolation across cities.

\begin{table}[!t]
\caption{Data division of our dataset (only image pairs that have ground-truth labels are split)}
\centering
\begin{tabular}{ccccc}
\hline
\textbf{City} & \textbf{Training} & \textbf{Validation} & \textbf{Testing} & \textbf{Total} \\ [2pt]
\hline
Shaoguan & 20 & 5 & 23 & 48 \\ [2pt]
Kunming & 200 & 3 & 50 & 253 \\ [2pt]
Yingde & 400 & 34 & 100 & 534 \\ [2pt]
Qichun & 600 & 80 & 200 & 880 \\ [2pt]
Wuhan & 0 & 0 & 20 & 20 \\ [2pt]
Hengyang & 0 & 0 & 22 & 22 \\ [2pt]
\textbf{Total} & \textbf{1220} & \textbf{122} & \textbf{415} & \textbf{1757}  \\ [2pt]
\hline
\end{tabular}
\label{tab3}
\end{table}

\section{Ground-truth Evaluation}
In this work, we assess the quality of our generated disparity maps. Two methods, namely image warping and SIFT matches, are used to evaluate the accuracy of the ground truth, which is similar to\cite{ref37} which uses the two methods to evaluate the US3D.

\subsection{Image Warping}
\begin{figure}[!t]
\centering
\includegraphics[width=3.45in]{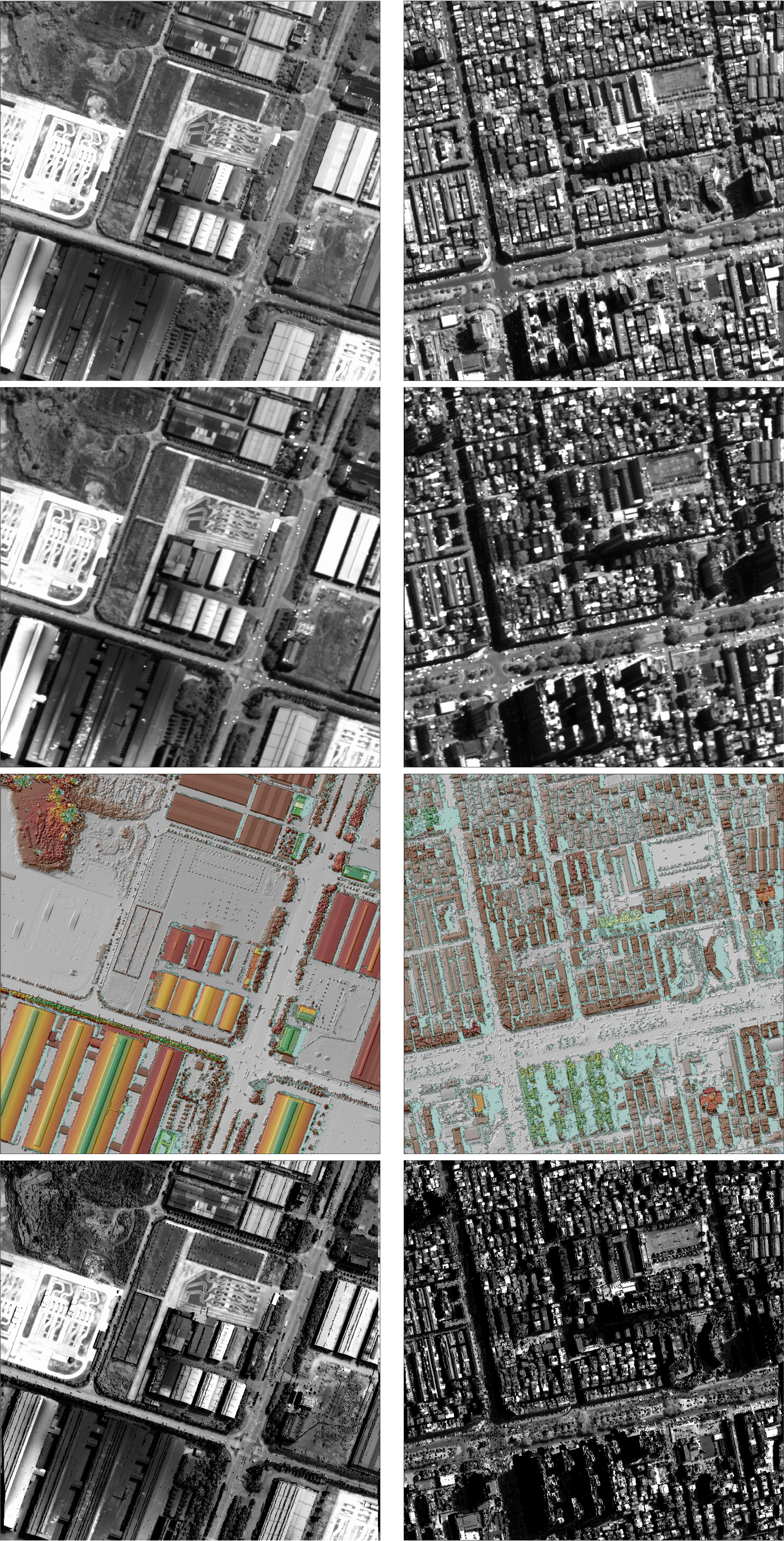}
\caption{Examples of inverse warping. From top to bottom: left images, right images, disparity maps (best viewed in color), and resampled right images.}
\label{fig9}
\end{figure}

The first method is based on inverse warping\cite{ref4}. We resample the right image in the left image geometry using the disparity map. Then, we compare the resampled right image to the left image. The difference observed between the two images results from radiometric differences between the original imageries, occlusions, resampling approximations, rectification errors, and disparity map errors. In our case, the disparity map errors are related to the disparity generation pipeline and the time discrepancy between the imageries and LiDAR point clouds. This method is mainly adopted for qualitative evaluation. Two examples of inverse warping are shown in Fig. \ref{fig9}.

\subsection{SIFT Matches}
The second method compares the disparity map with the horizontal distance of SIFT matches, which is adopted for quantitative evaluation. The disparity only exists in the horizontal direction between the left and right epipolar images. Due to matching mistakes, the generated key point pairs of SIFT may lead to vertical disparity. We do SIFT matching on the premise of row correspondence. Few key points are in integer positions, while pixel position is recorded as an integer. We need to find an integer position to evaluate the disparity map. We select key points in a local area with constant disparity on the left epipolar image and compare the difference between the disparity of keypoint pairs and the disparity of the integer position nearest to the key point. That means the selected key point has a half-pixel piecewise constant disparity in our assumption. Besides, to deal with time discrepancy between the point cloud and image, key points belonging to fewer time changes (e.g., streets, buildings) are more accurate in evaluating accuracy than vegetation. For images that contain many suburbs, we group them by vegetation and buildings according to if they include artificial facilities and assess the two categories separately.

Table \ref{tab4} shows the mean of the mean disparity errors and the mean of the mean absolute disparity errors. Kunming, Qichun, and Yingde have vegetation and buildings. Hengyang, Shaoguan, and Wuhan only have buildings. The mean of the mean disparity errors is finally worse in the vegetation class than in the building class. These vegetation class statistics could be worsened by difficult SIFT matching in texture-less or repetitive pattern regions. This vegetation class is more subjected to changes than the buildings class on this dataset. \cite{ref37} evaluated 49 pairs from the US3D dataset, the two indexes (for building) are reported as -0.20 and 0.89. Our dataset has a comparable mean of the mean disparity errors and a smaller mean of the mean absolute disparity errors to the US3D dataset. It indicates that the quality of our dataset is comparable to the same in the existing state-of-the-art stereo datasets like US3D.

\begin{table*}[!t]
\caption{Disparity errors (in pixels), computed using SIFT matches over epipolar image pairs and disparity maps}
\centering
\begin{tabular}{ccccc}
\hline
\textbf{City} & \textbf{Pair number} & \textbf{Matches number} & \textbf{Mean of the mean disparity errors} & \textbf{Mean of the mean absolute disparity errors} \\ [2pt]
\hline
Kunming (building) & 36 & 3576 & -0.69 & 0.48 \\ [2pt]
Kunming (vegetation) & 	216 & 54838 & -0.869 & 0.66 \\ [2pt]
Qichun (building) & 128 & 16643 & -0.11 & 0.51 \\ [2pt]
Qichun (vegetation) & 736 & 99457 & -0.5 & 0.79 \\ [2pt]
Yingde (building) & 90 & 13969 & 0.09 & 0.58 \\ [2pt]
Yingde (vegetation) & 438 & 63226 & 0.11 & 0.73 \\ [2pt]
Shaoguan (building) & 77 & 6744 & -0.24 & 0.42 \\ [2pt]
Wuhan (building) & 20 & 1776 & -0.37 & 0.49 \\ [2pt]
Hengyang (building) & 22 & 1139 & -0.03 & 0.32 \\ [2pt]
\hline
US3D (building) & 49 & 3750 & -0.20 & 0.89 \\ [2pt]
\hline
\end{tabular}
\label{tab4}
\end{table*}

\section{Experiment Evaluation}
To validate the utility of the presented dataset, we use it to evaluate some stereo matching methods and set their results as the baseline for our dataset. We first consider the conventional SGM\cite{ref6} algorithm, which is available in OpenCV and easy to implement with C++. Deep learning methods now significantly outperform most traditional stereo algorithms on public benchmarks, so we train and evaluate three CNN-based models with the WHU-Stereo dataset, namely StereoNet\cite{ref23}, PSM-Net\cite{ref22}, and HMSM-Net\cite{ref30}. We implement the networks by referring to their official open-source code with TensorFlow\cite{ref38} and Python, and all of the baseline implementations are provided along with our publicly available dataset.

All the CNN models are trained with the Adam optimizer ($\beta_{1}$=0.9, $\beta_{2}$=0.999) in an end-to-end manner. We standardize the images before they are input into the network for data preprocessing. To maintain the integrity of objects in these images, we directly feed the standardized image pairs with size $1024\times1024$ into the networks without cropping or resizing during training, and no data augmentation is adopted. We train the models from scratch up to 120 epochs. The initial learning rate is set to 0.001 and drops to half every 10 epochs as the training goes on. Only the best weights are saved during training, this is implementation-friendly in modern deep learning platforms. The disparity search range is set to [-128, 64], which covers almost all possible disparity values. Other hyper-parameters are set the same as the original papers have done. The deep learning models are trained using an RTX 3090 GPU on Ubuntu 20.04 OS and tested using a GTX 1080Ti GPU on Windows 10 OS, while the SGM method is implemented on Windows 10 OS.

Two metrics as proposed in\cite{ref26} are used for quantitative accuracy assessment, namely the average endpoint error (EPE) and the fraction of erroneous pixels (D1), which are expressed as:
\begin{equation}
EPE=\frac{1}{N} \sum_{k \in T}\left|\tilde{d}_{k}-\hat{d}_{k}\right|
\end{equation}
\begin{equation}
D1=\frac{1}{N} \sum_{k \in T}\left[\left|\tilde{d}_{k}-\hat{d}_{k}\right|>t\right]
\end{equation}
where $\tilde{d}_{k}$ and $\hat{d}_{k}$ represent the ground-truth and estimated disparity, $N$ and $T$ represent the number and the set of labeled pixels in the image, respectively, $t$ represents the threshold of erroneous disparity. EPE denotes the average absolute arithmetic difference between estimated and ground-truth disparities, and D1 represents the proportion of the erroneously estimated disparities to the total disparities (generally, an estimated disparity is regarded as erroneous if the absolute arithmetic difference between it and its ground-truth is larger than 3 pixels). The smaller the values of the two metrics, the better the stereo performance of the method is. Besides, we also evaluate the efficiency of different methods by computing the time required for processing an image pair.

\subsection{Traditional vs Deep Learning-based}
We first compare the traditional method with deep learning-based methods. For this purpose, we compute EPE and D1 for disparity maps estimated from all testing samples of the six cities and list the metrics in Table \ref{tab5} (these results can sever as the baseline of the dataset). As expected, all the deep learning-based methods significantly outperform SGM in accuracy. The recently published HMSM-Net, which is specially designed for stereo matching of satellite images, surpasses SGM by leap and bound while having nearly the same efficiency as SGM. And it is worth noting that Stereo-Net markedly exceeds SGM but only takes less than half the time of SGM.

\begin{table}[!t]
\caption{Accuracy comparison between SGM and deep learning-based methods}
\centering
\begin{tabular}{cccc}
\hline
\textbf{Method} & \textbf{EPE (pixel)} & \textbf{D1 (\%)} & \textbf{Time (ms)} \\ [2pt]
\hline
SGM & 4.889 & 50.79 & 506  \\ [2pt]
Stereo-Net & 2.453 & 25.12 & 238  \\ [2pt]
PSM-Net & 2.481 & 24.81 & 614  \\ [2pt]
HMSM-Net & 1.672 & 12.94 & 511  \\ [2pt]
\hline
\end{tabular}
\label{tab5}
\end{table}

\begin{figure*}[!t]
\centering
\includegraphics[width=7in]{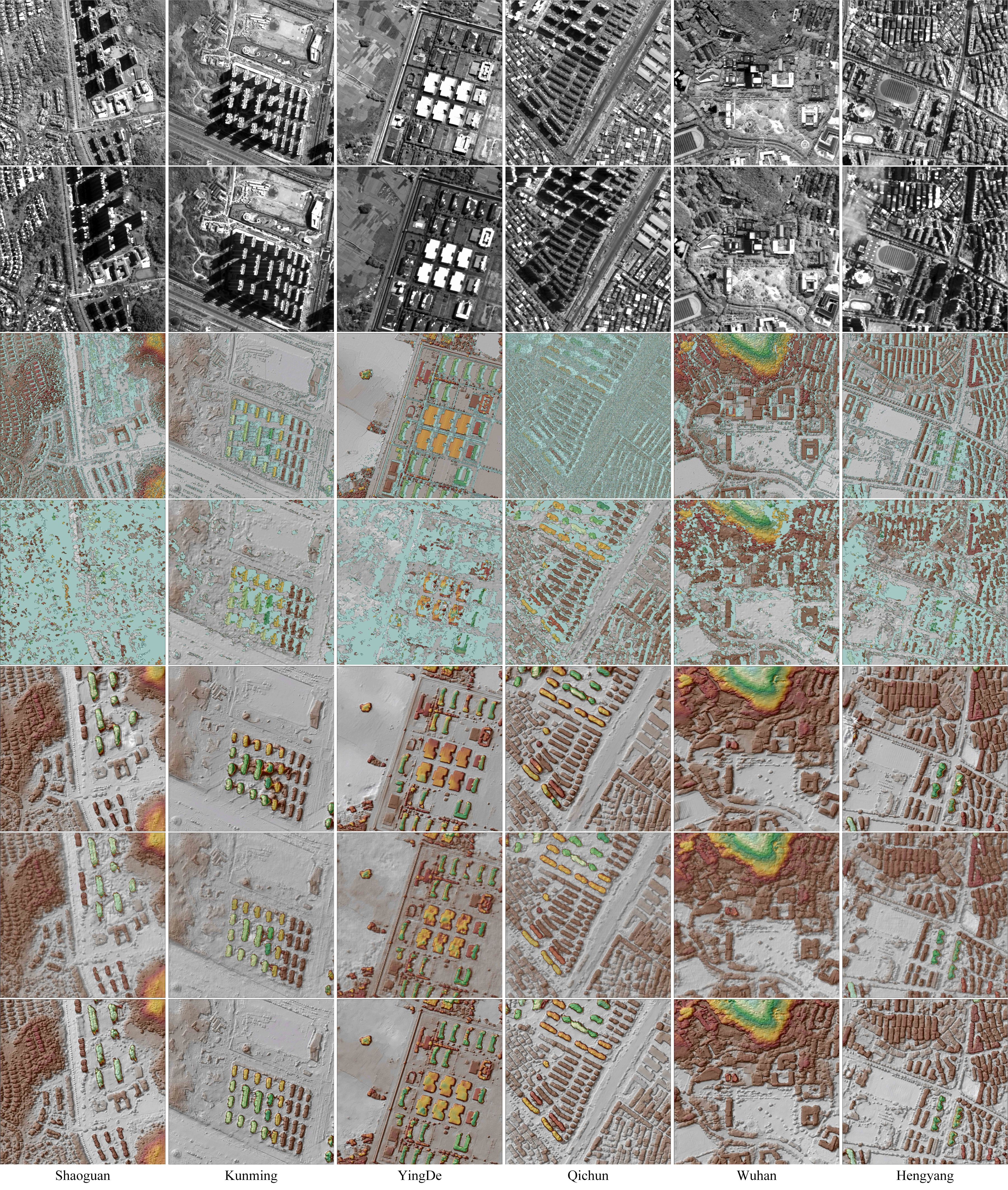}
\caption{Comparison of disparity maps estimated by different methods. From top to bottom: left image, right image, ground-truth, SGM, Stereo-Net, PSM-Net, and HMSM-Net.}
\label{fig10}
\end{figure*}

We also list some disparity maps predicted by these methods in Fig. \ref{fig10}. It can be seen that the deep learning-based methods generate much more satisfying estimated disparities in challenging regions, e.g., the repetitive vegetation, texture-less ground, and buildings that cause occlusion. Besides, there exist quite a few invalid values in the disparity maps generated by SGM, while deep learning-based models can infer a valid disparity for every pixel. And interestingly, deep learning methods are able to predict reasonable disparities even when there are thin clouds covering parts of the image (see the images from Hengyang). In summary, deep learning methods can easily outperform the traditional SGM when models are properly trained, and they can be well trained on our dataset.

\subsection{Generalization and Extrapolation of Deep Learning Models}
Deep neural networks have a vast amount of trainable parameters, which gives them the capacity to adapt to intricate details of training data distributions such as specific scene styles or common objects but also entails the danger of overfitting to local peculiarities that may be at odds with the goal to learn a generic prior valid for disparity estimation in unseen locations or even different, distant cities\cite{ref3}. In the following, we investigate the models’ geographical generalization \textit{(i)} within the same cities and \textit{(ii)} extrapolation across cities.

For this purpose, we separately calculate EPE and D1 for disparity maps estimated from testing samples of each city, as reported in Table \ref{tab6}. The models are trained using training sets from Shaoguan, Kunming, Yingde, and Qichun, results from these cities are used to analyze models’ ability of geographical generalization within a city, while that from Wuhan and Hengyang are used to analyze models’ ability of geographical extrapolation across cities. As depicted in the table, the deep learning models achieve high accuracy while testing images from the same cities, but the accuracy drops by a certain margin when testing unseen images from other cities. This indicates the models have a good ability of generalization within a city but extrapolation across cities needs to be improved. Nevertheless, deep learning models still perform better than the traditional SGM.

\begin{table*}[!t]
\caption{Accuracy of disparity maps estimated from testing samples of different cities}
\centering
\begin{tabular}{ccccccccc}
\hline
Method & \multicolumn{2}{c}{SGM} & \multicolumn{2}{c}{Stereo-Net} & \multicolumn{2}{c}{PSM-Net} & \multicolumn{2}{c}{HMSM-Net} \\ [2pt]
\cline{2-9}
City & EPE (pixel) & D1 (\%) & EPE (pixel) & D1 (\%) & EPE (pixel) & D1 (\%) & EPE (pixel) & D1 (\%) \\ [2pt]
\hline
Shaoguan & 7.862 & 63.19 & 2.660 & 24.58 & 2.514 & 21.45 & 2.091 & 16.94 \\ [2pt]
Kunming & 3.065 & 37.50 & 1.225 & 6.09 & 1.181 & 5.77 & 1.017 & 4.44 \\ [2pt]
Yingde & 5.497 & 47.22 & 1.821 & 14.69 & 1.993 & 15.09 & 1.436 & 9.86 \\ [2pt]
Qichun & 4.484 & 54.19 & 2.801 & 34.63 & 2.953 & 37.04 & 1.596 & 13.65 \\ [2pt]
\hline
Wuhan & 6.790 & 58.65 & 4.836 & 47.83 & 4.105 & 36.49 & 3.905 & 35.66 \\ [2pt]
Hengyang & 9.081 & 61.82 & 4.375 & 43.50 & 3.761 & 31.23 & 2.914 & 23.43 \\ [2pt]
\hline
\end{tabular}
\label{tab6}
\end{table*}

\subsection{Transfer Learning Investigation}
\begin{figure}[!t]
\centering
\includegraphics[width=3.45in]{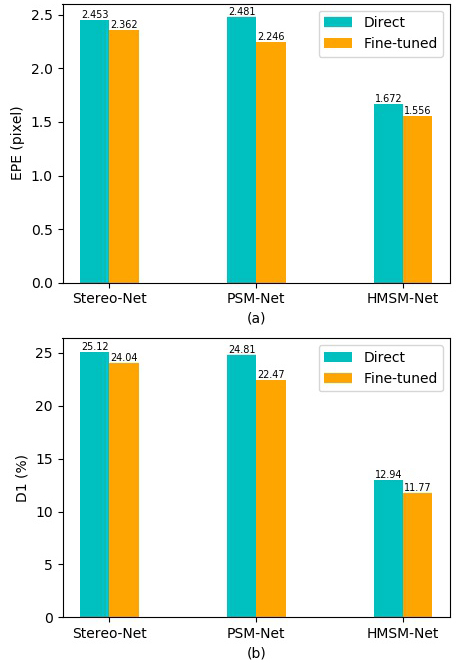}
\caption{Accuracy comparison between directly trained models and fine-tuned models.}
\label{fig11}
\end{figure}

Since the US3D dataset provides large-scale labeled data, we investigate whether a transfer learning strategy with fine-tuning can improve the models’ generalization and extrapolation ability. For this purpose, we first train these models on the US3D dataset for 30 epochs, then use the pre-trained weights to initialize the networks and train them on the training sets listed in Table \ref{tab3} for another 120 epochs. Fig. \ref{fig11} showcases the overall disparity accuracy comparison between directly trained models and fine-tuned models. Table \ref{tab7} showcases the separate disparity accuracy of each city. It can be observed that the fine-tuned models’ overall performance and performance in almost all cities have been improved although the images of the two datasets cover cities from two countries that have dramatically different landforms, indicating the transfer learning strategy works up to a point. Therefore, it might be a good choice to utilize available pre-trained models. One thing to note is that Stereo-Net’s performance on Kunming shows a slight decrease, thus how to maximize the benefit of transfer learning still needs further study.

\begin{table*}[!t]
\caption{Performance of fine-tuned models in different cities}
\centering
\begin{tabular}{ccccccc}
\hline
Method & \multicolumn{2}{c}{Stereo-Net} & \multicolumn{2}{c}{PSM-Net} & \multicolumn{2}{c}{HMSM-Net} \\ [2pt]
\cline{2-7}
City & EPE (pixel) & D1 (\%) & EPE (pixel) & D1 (\%) & EPE (pixel) & D1 (\%) \\ [2pt]
\hline
Shaoguan & 2.467 & 23.26 & 2.366 & 20.61 & 1.928 & 15.67 \\ [2pt]
Kunming & 1.245 & 6.19 & 1.091 & 4.86 & 0.996 & 5.05 \\ [2pt]
Yingde & 1.758 & 13.52 & 1.481 & 9.96 & 1.351 & 8.72 \\ [2pt]
Qichun & 2.760 & 34.07 & 2.817 & 35.05 & 1.483 & 12.09 \\ [2pt]
\hline
Wuhan & 4.346 & 42.88 & 4.025 & 34.92 & 3.718 & 34.51 \\ [2pt]
Hengyang & 3.885 & 39.18 & 3.553 & 34.85 & 2.539 & 19.62 \\ [2pt]
\hline
\end{tabular}
\label{tab7}
\end{table*}

\subsection{Potential Application}
\begin{figure*}[!t]
\centering
\includegraphics[width=7in]{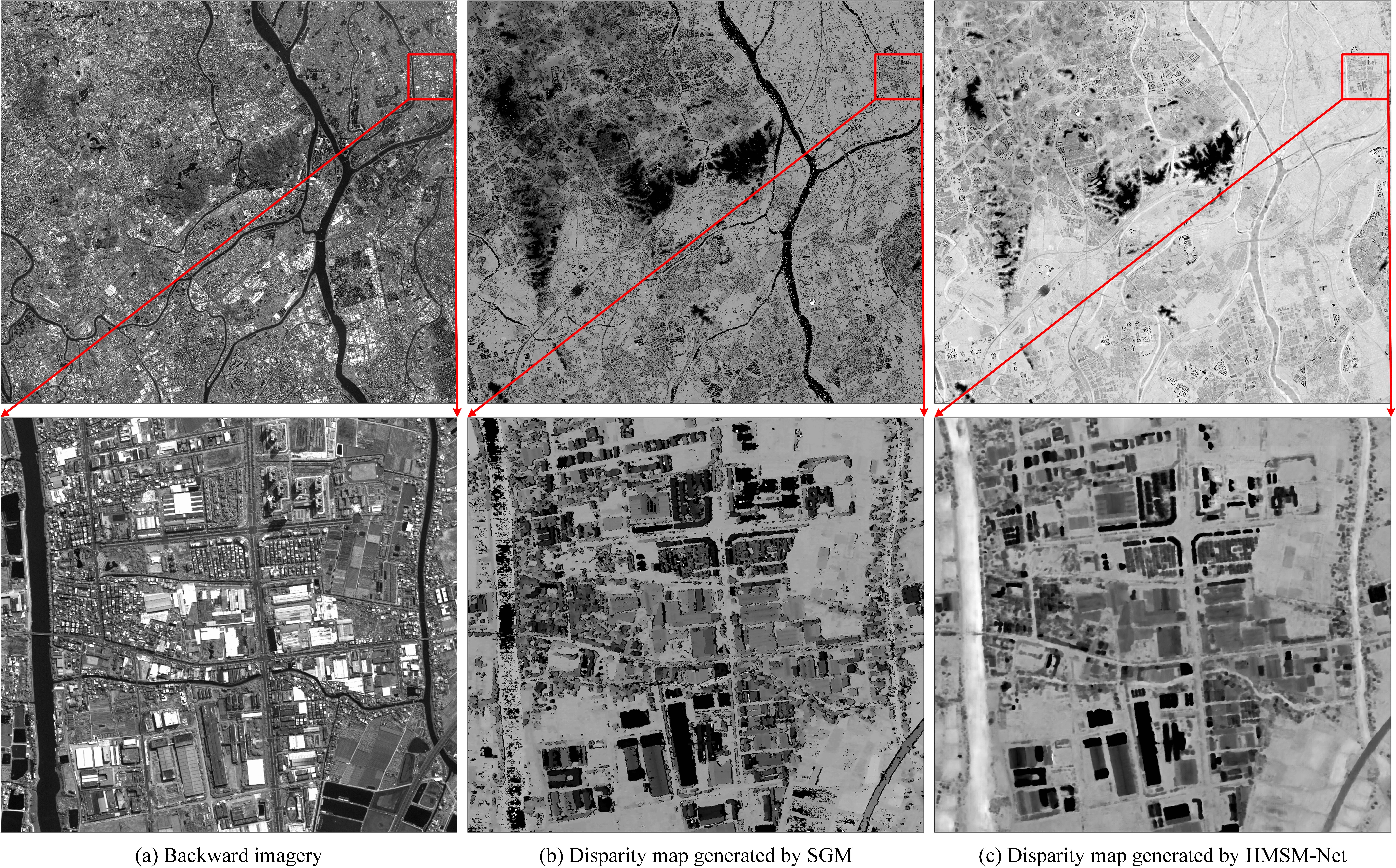}
\caption{Experimental result of disparity estimation of large-scale high-resolution satellite imagery.}
\label{fig12}
\end{figure*}

The above experiments are conducted on satellite image patches, in this work, we explore the potential application of a well-trained model on the stereo matching of satellite imagery with large size. We use all the samples of WHU-Stereo to train an HMSM-Net model and apply it to a pair of epipolar rectified GF-7 imageries (taken from Guangzhou City) that have a size larger than $34000\times30000$. To fit in with the GPU memory, we crop the imageries into pieces and estimate a disparity map for each piece, then the disparity maps are merged into a big disparity map aligned with the satellite imagery. The result is shown in Fig. \ref{fig12}. For comparison, we also adopt SGM to generate a disparity map. In general, both disparity maps generated by SGM and HMSM-Net can reflect changes in the height of the ground, but there exists more noise in the disparity map generated by SGM. In detail, HMSM-Net outputs disparities of higher quality, for example, disparities in the river are smoother and buildings are better discriminated. The comparison implies that a well-designed deep learning model has great potential in remote sensing imagery, as long as it is trained on a good dataset.

\section{Conclusion}
A large-scale, accurate, and publicly available dataset plays an indispensable role in promoting the development of deep learning techniques in remote sensing applications. In this paper, we provide the WHU-Stereo dataset, which is expected to contribute to developing and evaluating novel methods for stereo matching of satellite images. This dataset is produced using GaoFen-7 satellite imagery and airborne LiDAR point cloud and it is the first stereo matching dataset made from high-resolution imagery of Chinese cities, which is a good complement to existing databases. The dataset has been thoroughly evaluated and proved to be comparable to existing state-of-the-art datasets. In the experiment, we demonstrated the usage of the dataset, including evaluating the performance of recent studies in stereo matching and investigating transfer learning. The results demonstrate that a deep learning model easily outperforms the traditional SGM algorithm when it is properly trained and a transfer learning strategy helps to improve the performance of a model, and our dataset can sever as a challenging benchmark for evaluating a CNN-based model’s ability of geographical generalization within a city and extrapolation across cities. Besides, we demonstrate the potential application of a deep learning model and our dataset on the stereo matching of large-scale high-resolution satellite imagery.

Based on the current WHU-Stereo dataset, additional work is ongoing. We’re trying to collect more LiDAR data and remote sensing images, including both satellite and aerial imagery, to further expand the scale of our dataset. Moreover, development is in work to create thematic datasets, such as building extraction and building height estimation. And we’re interested in exploring more powerful deep learning-based stereo matching methods for remote sensing imagery as well.

\section*{Acknowledgments}
The authors would like to thank the China Centre for Resources Satellite Data and Application for providing the satellite imageries, and the developers in the TensorFlow community for their open-source deep learning projects, and the researchers in the stereo matching field for their open-source codes. The authors would also like to express their gratitude to the editiors and reviewers for their constructive and helpful comments.

\vfill

\end{document}